\documentclass[runningheads]{llncs}
\usepackage[T1]{fontenc}
\usepackage{graphicx}
\usepackage{hyperref}
\usepackage{color}
\usepackage{amsmath}
\usepackage{amsfonts}
\usepackage[dvipsnames]{xcolor}

\begin{document}
\title{Acrobotics: A Generalist Approach to Quadrupedal Robots' Parkour}
%
%
\author{Guillaume Gagn\'e-Labelle \and
Vassil Atanassov \and
Ioannis Havoutis}
\authorrunning{Guillaume Gagn\'e-Labelle et al.}
%
\institute{Dynamic Robot Systems Group (DRS), University of Oxford \\
\email{\{guillaume.gagne-labelle, vassil.atanassov, havoutis\}@oxfordrobotics.institute}}
\maketitle              
\begin{abstract}
Climbing, crouching, bridging gaps, and walking up stairs are just a few of the advantages that quadruped robots have over wheeled robots, making them more suitable for navigating rough and unstructured terrain. However, executing such manoeuvres requires precise temporal coordination and complex agent-environment interactions. Moreover, legged locomotion is inherently more prone to slippage and tripping, and the classical approach of modeling such cases to design a robust controller thus quickly becomes impractical. In contrast, reinforcement learning offers a compelling solution by enabling optimal control through trial and error. We present a generalist reinforcement learning algorithm for quadrupedal agents in dynamic motion scenarios. The learned policy rivals state-of-the-art specialist policies trained using a mixture of experts approach, while using only 25\% as many agents during training. Our experiments also highlight the key components of the generalist locomotion policy and the primary factors contributing to its success. Supplementary material and video can be found \href{https://drive.google.com/drive/folders/18h25azbCFfPF4fhSsRfxKrnZo3dPKs_j?usp=sharing}{ here}\footnote{\url{https://drive.google.com/drive/folders/18h25azbCFfPF4fhSsRfxKrnZo3dPKs_j?usp=sharing}}.

\end{abstract}
\section{Introduction}\label{sec:introduction}

Within the area of reinforcement learning, there are various methods for controlling quadruped robots. In the context of dynamic locomotion, the mixture of experts approach is nowadays considered the state of the art \cite{anymal_parkour,specialists_policy_humanoid,barkour,robot_parkour}. The technique consists of training one machine learning model per targeted ability, for example, walking, climbing, trotting, etc., and then training an upstream model to select the appropriate skill at any given moment. From an explainability perspective, decoupling the learning process into sub-parts allows an easier analysis of the derived policy. Additionally, splitting the problem into smaller components simplifies the overall solution and the engineering choices required, since reinforcement learning (RL) algorithms are known to be mroe sensitive to design choices than other machine learning paradigms \cite{Sutton1998,unstable_rl}.

Nevertheless, the mixture of experts technique is not devoid from substantial drawbacks. Three of them are particularly worth addressing. First, the mixture of experts does not make optimal usage of the neural networks' parameters. For example, the limb coordination required to jump on a box or to leap over a gap is very similar. Therefore, some information accumulated in each specialized network parameter is repeated. Decoupling these actions in separated models \cite{robot_parkour} can lead to a sub-optimal architecture where a larger than necessary set of parameters is needed to achieve the same performance.

Another downside of the mixture of experts approach is that the algorithm pipeline can be overly complex and difficult to optimize and improve. This is especially true for RL problems since often other inference models (e.g. critics) are involved during the training phase. Furthermore, for legged locomotion, it is already common to split the problem into a planning module (to generate trajectories) and another path-following module or controller (to steer the robot towards the generated path) \cite{locomotion_end_to_end}. In this setting where networks are inevitably entangled in the algorithm, splitting the task into multiple specialized networks becomes even more challenging.

Lastly, the mixture of experts introduces inductive biases by defining what constitutes a specific skill and how information flows, which can lead to sudden, non-smooth, transitions between policies \cite{barkour}. As argued in Richard Sutton’s The Bitter Lesson \cite{the_bitter_lesson}, history shows that generalist methods often outperform approaches based on domain knowledge, highlighting the need for better algorithms rather than human-driven designs.

These observations motivate the work presented throughout this paper. We propose a generalist policy with a less convoluted algorithm pipeline. Our goal is to analyze the algorithm thoroughly by conducting ablation studies and implementing prominent generalist policies' ideas.

\subsection{Approach}

We develop a generalist policy that uses deep reinforcement learning (DRL) to control quadruped robots. In this context, a generalist policy, as opposed to a specialist one, is a single unified policy that can tackle all presented obstacles without the need for ad hoc switching between different specialized policies when encountering specific obstacles.

Despite their promising potential for generalization, generalist policies have multiple weaknesses. In particular, they tend to be less stable and more sensitive to design choices \cite{campanaro2023a,barkour,anymal_parkour}. Consequently, the goal of this work is not to outperform the state-of-the-art with arbitrary engineering choices. Instead, this paper aims to study and analyze the foundational elements necessary for the convergence and optimal performance of generalist agents while focusing on simplicity and stability. These findings aim to offer valuable insights for developing larger and more powerful generalist agents by leveraging greater computational resources and integrating the best-performing methods from this work into future projects.

\subsection{Contribution}
Our contribution has two parts. First, we propose a simple yet competitive generalist algorithm for training quadruped robots. Our pipeline requires fewer computational resources than concurrent methods, the proposed algorithm is simple and easy to implement, and the performance is comparable to state-of-the-art specialist policies for agile locomotion. Second, we present a rigorous and thorough ablation study of the main implementation components. This analysis of the algorithm helps identify the critical design choices and ideas that are helpful for the convergence of reputedly unstable RL algorithms. This can greatly help researchers in designing learning-based locomotion approaches and develop systems significantly faster. 

\section{Related Work}\label{sec:related_work}

\textbf{Legged Locomotion:} 
Addressing the challenge of legged locomotion has long relied on complex model-based methods \cite{industrial_offshore_inspection,autonomous_spot} that only offer partial solutions. 
Recently, DRL has shown to be a compelling alternative and demonstrated state-of-the-art results on both proprioceptive \cite{hwangbo_learning_2019,lee_learning_2020,margolis_walk_2023,gangapurwala_learning_2023,10758728} and perceptive \cite{miki_learning_2022,campanaro2024a,gangapurwala_rloc_2022,miki_learning_2024} locomotion on difficult terrain. Many recent methods follow a ``privileged learning'' approach \cite{chen_learning_2020}, where a ``teacher'' agent is first trained with privileged information (e.g., environment parameters, mass and ground friction), and then distilled into a ``student'' network that is trained to replicate the teacher's actions given non-privileged observations \cite{miki_learning_2022,kumar_rma_2021}.
%

\textbf{Robot Parkour:} Given the success of DRL for locomotive policies, recent research has focused on demonstrating ever-more agile parkour-like behaviors \cite{extreme-parkour,robot_parkour,barkour}. As these skills are more challenging to learn, prior work often trains a discrete set of specialist policies for different parkour sub-tasks, and either distills them into one generalist policy \cite{robot_parkour,barkour} or trains a high-level skill selector \cite{anymal_parkour}. This adds a significant overhead, partially due to designing and tuning the rewards for each specialist policy, and due to the computational cost of the distillation step, especially for perceptive controllers that requires depth camera images.



\section{Method}\label{sec:method}

\begin{figure}[htbp]
    \centering
    \includegraphics[width=1.0\linewidth]{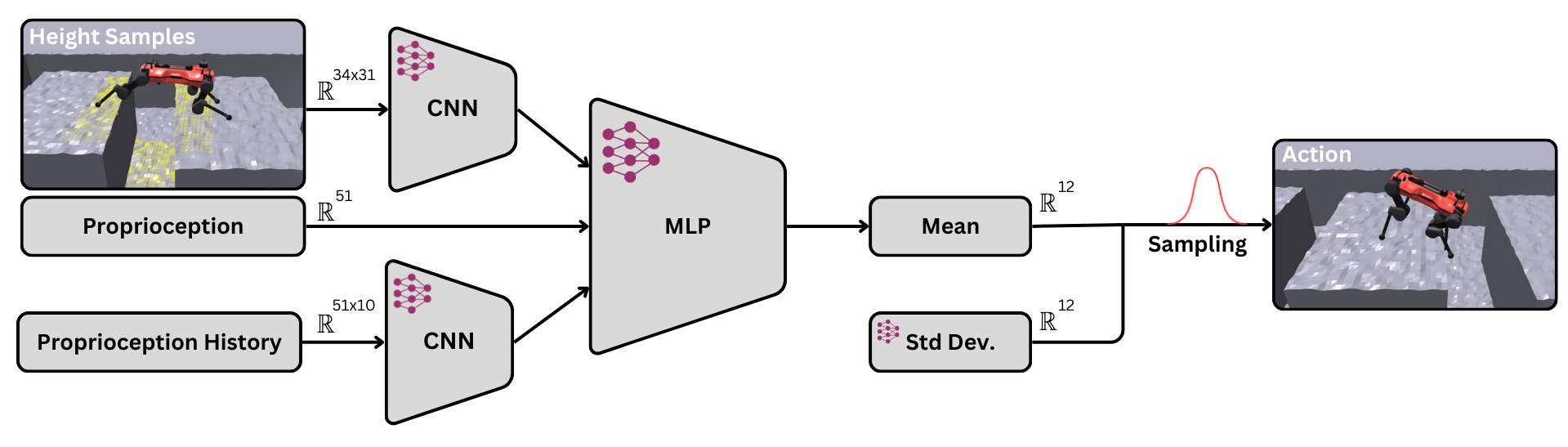}
    \caption{Training overview. The height samples and observation history are encoded through a convolutional neural network to reduce their dimensionality before being concatenated and given to the main actor.}
    \label{fig:architecture}
\end{figure}

Our approach uses a generalist policy trained on Isaac Gym \cite{makoviychuk2021isaacgymhighperformance} via Proximal Policy Optimization (PPO) \cite{ppo}. The training phase requires about four hours on an NVIDIA RTX 3060 GPU. The critic network is modeled with a multilayer perceptron with three hidden layers of size 512, 256 and 128 that outputs a scalar. The policy $\pi$ uses the same architecture as the critic, but the output is of size 12.

Each value in the policy output corresponds to the mean of a Gaussian distribution for each joint from which the action is sampled at every timestep. The standard deviation of these distributions is modeled with a single parameter updated via gradient descent on the RL objective. The standard deviation is limited to a maximum value of 1.0 to prevent an unrestrained increase in the parameter. This adjustment prevents pathological snowballing effects of the standard deviation observed in \cite{extreme-parkour} and stabilizes the overall algorithm.

The reward function structure is available in the supplementary material. Notably, as encouraging steady large steps has been a successful approach in other works \cite{walk_in_minutes,agile_and_dynamic_motor_skills}, there is a heavy penalty for having all four limbs in the air at the same time and a reward for promoting a trotting gait with long stride only in the flat environment.

For the agent to learn a stable gait before engaging in parkour, a small pretraining phase of 1k iterations on a flat terrain is introduced. It takes about 5k total iterations for an agent to learn to behave stably in every environment. The training is extended to 20k iterations before assessing the performance.

The Barkour benchmark \cite{barkour} introduced standardized metrics to evaluate the performance of quadruped robots through an obstacle course inspired by dog agility competitions. The course incorporates elevated pause tables, weave poles, a 30° A-frame, and a jump board. The scoring system considers both the completion time and penalties. Unlike Barkour, the goal of this work is not to maximize agility directly but rather to overcome obstacles at a given steady pace. Our work thus adapts the benchmark by adjusting the difficulty to one that is challenging yet achievable for the ANYmal robot: that is, about 70\% of the maximal difficulty. The proposed adaptation of the score $R_{\text{adapted}}$ is
\[R_{\text{adapted}} = 1.0 - |t_\text{run} - \mathbb{E}(t_\text{command})| * 0.01 - \text{penalties}\]
where $t_{run}$ is the time required to complete the 18m of the parkour course and $\mathbb{E}(t_{command})$ is the expected time to complete it given the input of an arbitrary velocity command. A penalty of $-0.1$ is added to the final score every time the robot trips or fails an obstacle. An agent unable to complete the course receives a score of 0. The reported adapted Barkour score is an average over 30 runs evaluated manually for each experiment.

\subsection{Input}

The policy $\pi$ and the critic share the same input, which is a concatenation of three components: the proprioception, the elevation map and a history of previous proprioceptive states. Both the proprioception and the elevation map are provided by the ANYmal robot framework, in both simulation and on the real robot platform.

\textbf{Proprioception.} This part of the input is a 51-dimensional concatenation of the following:
base linear velocity $\mathbf{v} \in \mathcal{R}^3$, base angular velocity $\mathbf{\omega} \in \mathcal{R}^3$, roll and pitch angles $\mathbf{\alpha} \in \mathcal{R}^2$, oracle heading $\mathbf{\psi} \in \mathcal{R}^2$, velocity command $v_{cmd} \in \mathcal{R}^1$, joint position $\mathbf{q} \in \mathcal{R}^{12}$, joint velocity $\mathbf{\dot{q}} \in \mathcal{R}^{12}$, previous action $\mathbf{a} \in \mathcal{R}^{12}$, and feet contact state $\mathbf{c} \in \{ 0, 1\}^4$.
The oracle heading represents the yaw angle between the heading of the robot and the target. We provide the following two targets to achieve better performance. The velocity command is sampled in the range $[0.4,0.8]{\text{ m/s}}$. The position of each joint is given in radians from the respective articulation, where the origin corresponds to the nominal position of the robot.

\textbf{Elevation Map Encoding.} The elevation map corresponds to 1054 evenly spaced height samples. The samples come from $34 \times 31$ points from a $1.6 \times 1 \text{m}^2$ region under the robot. The region is slightly shifted forward from the center of the robot. The height samples are the only available external perception of the robot in the algorithm. To leverage the spatial coherence of the samples, they are encoded into a vector of 32 dimensions by a convolutional neural network (CNN).

\textbf{History of Proprioceptive States.} The history is a buffer of the ten previous proprioceptive states of the robot. The $51 \times 10$ dimensional input is first reduced to a $30 \times 10$ dimensional matrix by the means of a linear layer before being encoded with a CNN sliding in the temporal dimension. The output is encoded in a 20-dimensional vector.

\subsection{Training Environments}\label{sec:training_env}
\begin{figure}[h!]
    \centering
    \includegraphics[width=1.0\linewidth]{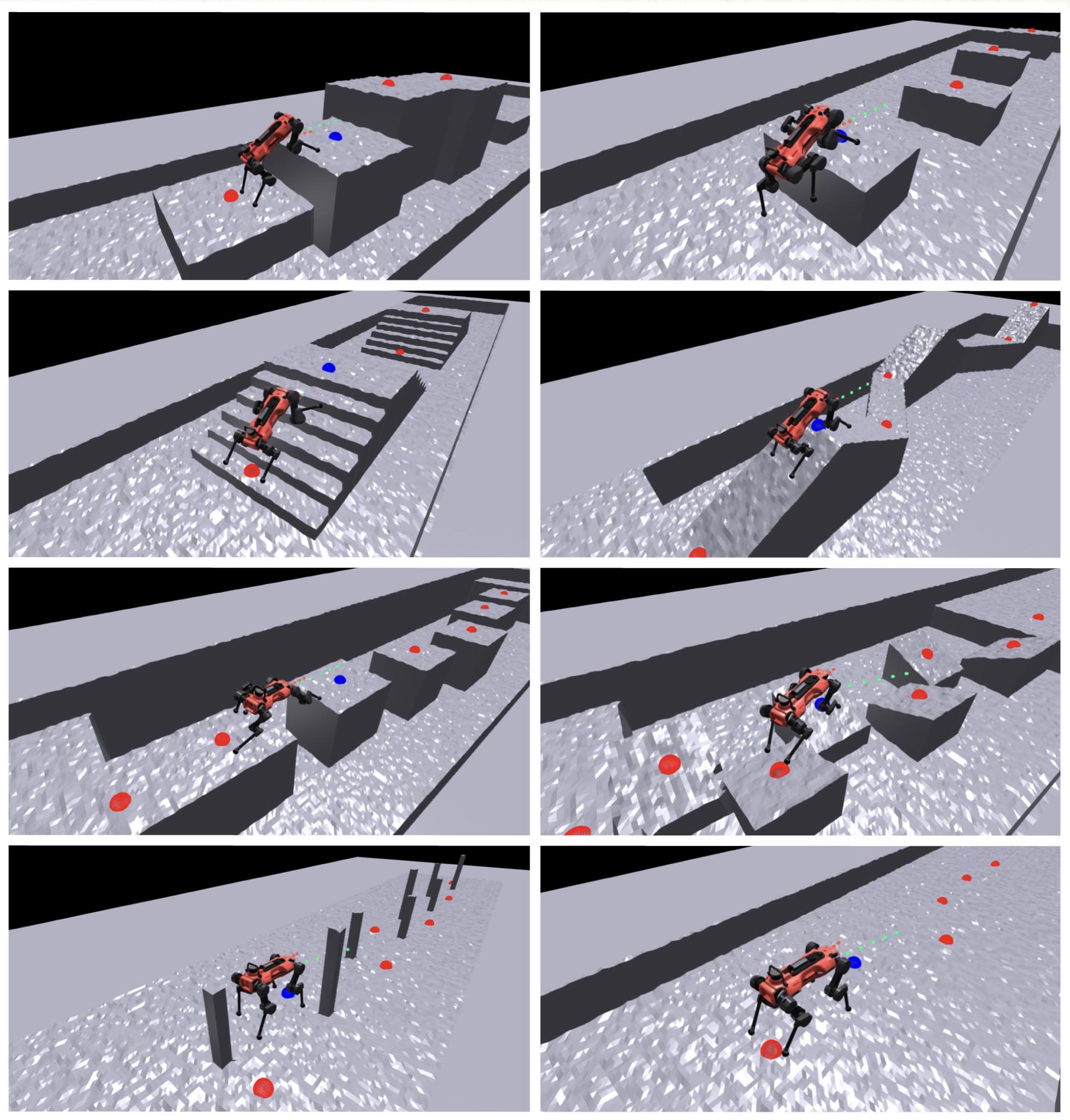}
    \caption{IsaacGym simulation setup of the eight environments used during the training phase represented at 75\% of their maximal difficulty.}
    \label{fig:training_envs}
\end{figure}

1024 ANYmal agents are trained in parallel in a variety of eight categories of different environments presented in Figure \ref{fig:training_envs}. Each terrain is modelled as an $18 \times 4 \text{ m}^2$ area that the agent has to overcome. The flat terrain is used mainly during the pretraining phase, but also sparsely during the training phase to avoid forgetting after the distribution shift from pretraining to training \cite{forgetting_deep_learning}. The difficulty ranges are reported in Table \ref{tab:obstacle_diff_range}.

\begin{table}
    \centering
    \caption{Difficulty range of the training obstacles}
    \begin{tabular}{cccccc}
    \hline
        \textbf{Obstacle} & Steps  & Boxes & \multicolumn{2}{c}{Stairs} & Gaps\\
    \hline
    \hline
        \textbf{Parameter} & height (m)    & height (m)  & tread (m) & riser (m) & gap size (m)\\
        \textbf{Difficulty} & $[0.1, 0.8]$ & $[0.1, 1.0]$ & $[0.3, 0.5]$ & $[0.05, 0.25]$ & $[0.1, 1.0]$ \\
    \hline
                        & \multicolumn{2}{c}{Inclined Boxes} & Slopes & Weave Poles\\
    \hline
    \hline
        \textbf{Parameter} & tilt (\textdegree) & stone length (m) & angle (\textdegree) & spreading (m)\\
        \textbf{Difficulty}& $[0, 50]$ & $[0.9, 1.8]$ & $[10, 30]$ & $[0.1, 0.7]$ \\
    \hline
    \end{tabular}
    
    \label{tab:obstacle_diff_range}
\end{table}

The difficulties of the terrains are modeled as a discrete curriculum of 20 levels. Let $[\alpha, \beta]$ be a difficulty range for an arbitrary obstacle (as shown in Table \ref{tab:obstacle_diff_range}), the difficulty $\mathcal{D}$ of level $l$ is calculated as \( \mathcal{D} = \alpha + l \cdot (\beta - \alpha)/20\). Once an agent has completed the last level, it is reassigned to a random level of the same category.

To get promoted to a more challenging terrain, an expected distance is computed from the velocity command and a time window of 20 seconds. The robot gets promoted to a harder level if it crosses more than 80\% of this expected distance and gets demoted if it doesn't traverse more than 40\%.

\section{Results \& Discussion}\label{sec:results_and_discussion}

Table \ref{tab:prediction_interval} displays the 90\% prediction intervals for the success rate, adapted Barkour score, and average sum of squared torques \footnote{The average sum of squared torques on successful agents serves as a proxy for energy loss --- which is itself a proxy for natural locomotion of quadruped animals \cite{locomotion_end_to_end}} after 5k learning iterations. For comparison purposes, Table \ref{tab:prediction_interval} also displays the performance after 20k learning iterations of the algorithm as well as the performance of the first phase (``teacher'' training) of the Extreme Parkour with Legged Robots algorithm \cite{extreme-parkour} after the same number of iterations.

\begin{table}[h!]
    \centering
    \caption{Performance of the main algorithm on the adapted Barkour environment.}
    \begin{tabular}{lcccc}
        \hline
        \textbf{Algorithm} & Completion (\%) & Barkour Score & Avg. Torque ($N^2 m^2$)\\
        \hline
        \hline
            Base model - 5k   &  $[87.26, 100+]$ & $[0.73, 0.97]$ & $[14'526, 15'990]$ \\
        \hline
            Base model - 20k  & 99.61 & 0.90 & 13'989 \\
            Ex. Parkour - 20k & $0.00$ & $0.00$ & N.A. \\
        \hline
    \end{tabular}\label{tab:prediction_interval}
\end{table} 

After 20k learning iterations, the completion percentage is not significantly different from the model at 5k iterations as it falls into the prediction interval, but this is likely due to the evaluation environment being slightly too easy for both models in that particular case. The 20k model provides better completion percentage than every individual 5k model \footnote{The exact results of
the four models from which the prediction intervals are derived are presented
in the supplementary material: \href{https://drive.google.com/drive/folders/18h25azbCFfPF4fhSsRfxKrnZo3dPKs_j?usp=sharing}{\footnotesize \url{https://drive.google.com/drive/folders/18h25azbCFfPF4fhSsRfxKrnZo3dPKs\_j?usp=sharing}}.}. Moreover, the average sum of squared torques is significantly below the prediction interval, indicating improvement in terms of unnecessary movements. After 20k iterations, the agent learns to gallop over gaps rather than crossing them one foot at a time, and the gait is generally much smoother.

The adapted Barkour score is 0.90, where a single penalty was observed over thirty runs when the robots struggled to climb on the ending table. The main factor in lowering the score is the speed of the robot. The agent speed consistently exceeds the velocity command by $(41 \pm 16)\%$ on average. This behavior is most probably a consequence of the goal-tracking reward. Indeed, as can be seen in the provided supplementary material, the agent can ensure a maximal goal-tracking reward by aiming almost directly at the next goal at a speed greater than the velocity command. Nonetheless, observation shows that the velocity command still affects the agent's speed, probably because crossing obstacles at lower speeds is generally easier. As for the agent trained with the Extreme Parkour with Legged Robot algorithm, although it is able to complete some of the parkour obstacles with low probability (see Figure \ref{fig:main_results}), completing all five obstacles consecutively proved too difficult and no successful run was observed after 30 attempts.

\subsection{Fundamental Skills}

\begin{figure}[h!]
    \centering
    \includegraphics[width=0.85\linewidth]{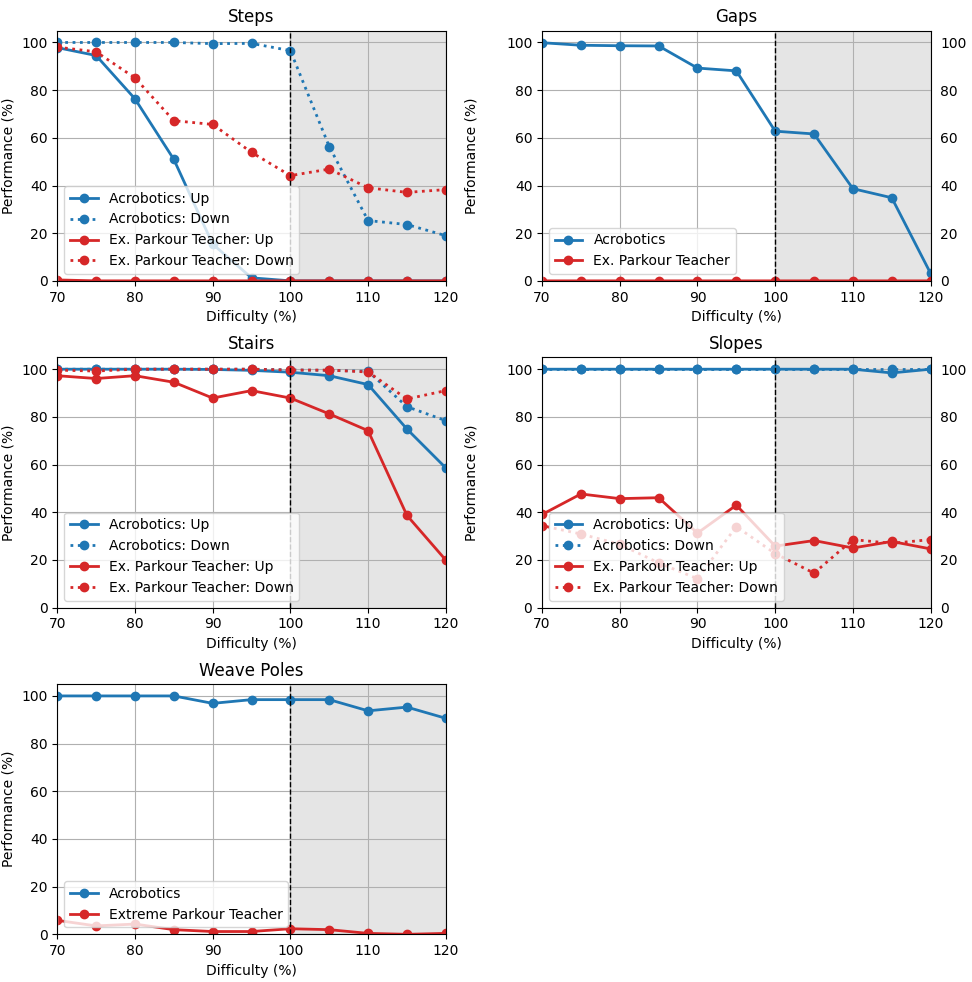}
    \caption{Performance of the algorithm on the basic skills developed through the training environments.}
    \label{fig:main_results}
\end{figure}

Figure \ref{fig:main_results} depicts the empirical probability of overcoming the obstacle for each skill. The robot achieves about 90\%+ performance on every single obstacle at 95\% of the training difficulty besides the climbing up skill. The performance is still excellent for obstacles up to approximately 75cm, corresponding to the robot's height to its hips. The drop in performance at this height is intuitive: when observing the policy for boxes above this threshold, it seems like the robot struggles to put its knee on the box to push itself up. Even when fully opening its hip, its knee is only about 75cm high.

As for gaps, there seems to be two limiting factors. The first one is the robot's reach, but the second one is the extent of the elevation map in front of the robot, which is roughly 90cm and where the performance seems to drop. Past that point, as the elevation map is the only exteroception of the agent, it is difficult to discern between an edge and a gap. Moreover, to bridge wide gaps, the robot benefits from linear momentum. This momentum is difficult to obtain at low velocities when the detection the other side of the gap happens late.

Without using any form of privileged information, Acrobotics consistently outperforms the teacher of the algorithm presented in Extreme Parkour \cite{extreme-parkour} when applied to ANYmal. A probable reason for this phenomenon is the lower joint velocity of the limbs of ANYmal compared to the A1 robots for which Cheng et al. designed and applied their algorithm to. As discussed in Section \ref{sec:method}, during the teacher's training phase, the average standard deviation of the actions increases over time, yielding almost random actions after an extensive training phase and preventing any meaningful learning process. By clipping the said standard deviation of each action, the exploration is limited, but this seems to be an advantageous trade-off in the long run.

At first glance, it appears in Figure \ref{fig:main_results} that Extreme Parkour outperforms Acrobotics in the 100\%+ difficulty regime when climbing down stairs and stepping down boxes. Observation shows that our algorithm tends to be more meticulous when approaching those obstacles and a clear inclination to keep the robot's feet on the ground is noticeable. This leads to a tendency to tip over when a step is too high or a staircase, too shallow. On the other hand, the behavior of the robot trained with the Extreme Parkour approach seems more impulsive: it tends to jump off or fall off the box or staircase. The robot sometimes ``survives'' the fall (it doesn't violate any termination condition), but such carelessness can limit real world deployment in practice.

\subsection{Ablation Study}

All the policies presented in this section are compared and evaluated after 5k iterations. In each table, a value that lies within the 90\% confidence prediction interval of the base model performance is presented in green: no statistically significant impact on performance is observed from the ablation. An utterly green row signifies that no evaluated metric is altered from the ablation; the component is superfluous to the algorithm's performance.

\subsubsection{Proprioception.} The results reported in Table \ref{tab:ablation_input} highlight that, with 90\% certainty, the algorithm extracted from the ablation of each part of the proprioception input comes from a different distribution than the one resulting from the default algorithm. In particular, the linear velocity of the base of the robot, the position of the robot's joints and the previous sampled action are unequivocally the most important components of the proprioception as the performance quickly deteriorates without these inputs. On the other hand, although necessary, the information provided by the angular velocity and the detection of the feet contacts with the ground seems to be less important pieces of information in proprioception. In both cases, the null hypothesis is rejected in favor of the alternative hypothesis: the ablation of the feet contacts detection has a statistically significant impact on the performance of the algorithm.

\begin{table}[h!]
    \centering
    \caption{Performance of the algorithm on the adapted Barkour environment after training with ablation of the input. An ablation is done by masking the according input with zeros in the input vector.}
    \begin{tabular}{lcccc}
        \hline
        \textbf{Ablation} & Completion (\%) & Barkour Score & Avg. Torque ($N^2 m^2$) \\
        \hline
        \hline
            Base Model         & $[87.26, 100+]$ & $[0.73, 0.97]$ & [14'526, 15'990] \\
        \hline
            Linear Vel. (3D)   &  9.77 & 0.06 & 19'197\\
            Angular Vel. (3D)  & 82.13 & 0.85 & 15'533\\
            Roll \& Pitch (2D) & 75.29 & 0.75 & 15'547\\
            Joints Pos. (12D)  &  0.00 & 0.00 &   N.A.\\
            Joints Vel. (12D)  & 76.86 & 0.60 & 16'007\\
            Prev. Action (12D) & 18.65 & 0.12 & 19'795\\
            Feet Contact (4D)  & 86.04 & 0.64 & 15'045\\
            \textcolor{Green}{History ($510$D)} & \textcolor{Green}{88.18} &\textcolor{Green}{0.80} & \textcolor{Green}{14'901}\\
        \hline\label{tab:ablation_input}
    \end{tabular}
\end{table}

\subsubsection{History of Proprioceptive States.} As per Table 3, the results from the proprioception history input seem to be nuanced. Generally, every evaluated metric falls inside the prediction interval and the null hypothesis can not be rejected. However, both the completion rate and the adapted Barkour score fall not only in the lower part of the prediction interval but are individually equal or worse than every reported metric of the individual models. Therefore, it might be that the impossibility of rejecting the status quo hypothesis is due to the small amount of under-trained default models rather than equal performances. In other words, only four default models were trained for only 5k iterations to build the prediction intervals. This creates uncertainty with respect to the performance that is reflected in the prediction intervals. Since the performance of the ablation rests in the lower part of the intervals, it is possible that the performance is affected but not quantifiable in this context.

\subsubsection{Reward Function.} Quantifying the role of specific terms in the reward function is not straightforward. Some of the terms aim not to increase the overall performance directly but rather to improve the policy's natural-looking behavior. Even though the average sum of squared torques tries to quantify precisely this, it is only a proxy for the minimization of energy which is itself only a proxy for natural-looking behavior. The conclusions extracted from this subsection thus provide valuable insights, but should be considered within the context of the outlined constraints.

\begin{table}[h!]
    \centering
    \caption{Performance of the algorithm on the adapted Barkour environment after training with ablation of the reward function.}
    \begin{tabular}{lcccc}
        \hline
        \textbf{Ablation} & Completion (\%) & Barkour Score & Avg. Torque ($N^2 m^2$) \\
        \hline
        \hline
            Base Model           & $[87.26, 100+]$ & $[0.73, 0.97]$ & [14'526, 15'990] \\
        \hline
              Base Angular Vel.  & 79.10 & 0.62 & 15'970\\
              Joint Acceleration & \textcolor{Green}{91.99} & \textcolor{Green}{0.81} & 16'677\\
              Collision          & 44.92 & 0.43 & 17'535\\
              Action Rate        & 0.20  & 0.00 & 16'136\\
              \textcolor{Green}{Torques Variation}  & \textcolor{Green}{89.45} & \textcolor{Green}{0.80} & \textcolor{Green}{14'996}\\
              Torques            & 1.17  & 0.00 & 31'588\\
              DOF Error          & 83.59 & 0.81 & 15'953 \\
              Stumbling          & \textcolor{Green}{92.77} & \textcolor{Green}{0.85} & 16'148\\
              Trotting           & 65.82 & 0.47 & 14'300\\
              Feet in Air        & 83.40 & 0.67 & 14'899\\
              Feet Air Time      & 60.16 & 0.60 & 14'314\\
        \hline\label{tab:ablation_reward_function}
    \end{tabular}
\end{table}
Table \ref{tab:ablation_reward_function} shows unequivocally that the roles of the torque and action rate penalty are crucial to the success of the derived policy. These two penalties are subtly different from one another. The action rate term penalizes the agent for taking actions that are widely different at two sequential timesteps. On the other hand, the torque is directly proportional to the angular acceleration of the joint. Therefore, the torque penalty constrains the robot into taking actions that limit the acceleration of its joints. This means, for example, that two subsequent actions that accelerate a limb in the same direction are greatly penalized by the torque penalty but not by the action rate penalty. Both terms seem to be critical to the functionality of the algorithm. 

Surprisingly, the penalty associated with torque variation does not significantly influence performance. The meaning of this penalty is very similar to the action rate: it penalizes the agent for taking subsequent actions that create widely different torques. A possible explanation of the negligible consequences of the ablation of the torque variation is the reward's coefficient, as it is one million times smaller than the action rate's coefficient.

For the same reason of being multiplied by a very small coefficient, the ablation of the joint acceleration penalty, with a reward coefficient only 2.5 times greater than the torque variation penalty, does not affect the performance. Nevertheless, the ablation of both the joint acceleration and torque variation has a statistically significant impact on the average sum of squared torques and might promote the natural-looking behavior of the agent.

\subsubsection{Pretraining, Standard Deviation Limit \& Batch Size.} This ablation study focuses on assessing the role of three higher-level design choices of the algorithm. It questions the necessity of the pretraining phase, which lasts 1k iterations, or 20\% of the total training duration. It also challenges the idea of introducing a maximal limit of 1.0 for the standard deviation of each action distribution and the importance of the batch size in the RL pipeline.

\begin{table}[h]
    \centering
    \caption{Performance of the algorithm on the adapted Barkour environment after training with ablation of high-level design choices.}
    \begin{tabular}{lcccc}
        \hline
        \textbf{Ablation} & Completion (\%) & Barkour Score & Avg. Torque ($N^2 m^2$) \\
        \hline
        \hline
            Base Model            & $[87.26, 100+]$ & $[0.73, 0.97]$ & [14'526, 15'990] \\
        \hline
            STD Clipping  & 63.96 & 0.61 & 24'852\\
            Pretraining & \textcolor{Green}{94.14} & \textcolor{Green}{0.88} & 16'007\\
            Batch Size 512 & 0.39 & 0.00 & 17'141\\
        \hline \label{tab:ablation_pt_std_batch}
    \end{tabular}
\end{table}

Table \ref{tab:ablation_pt_std_batch} shows that above all else, batch size is the most crucial design choice. Decreasing the batch size from 1024 to 512 implies that the policy encounters only half as many scenarios per learning iteration. When there are not enough agents at any given time in the environment, the reward signal is noisier and the variance hinders the learning process. More experiments were performed with even fewer agents, but the results are not reported as the performance completely deteriorated. It is reasonable to believe that increasing the number of agents by leveraging more computational power could significantly improve performance. 

The ablation of the standard deviation clipping also has a statistically significant impact on performance according to every metric. Potentially, as the standard deviation increases over training time, rather than promoting exploration, the actions become increasingly jittery and almost random. By limiting the standard deviation magnitude, some exploration is still possible, but the sampled actions are still meaningful. 

Finally, the ablation of the pretraining phase does not provide a significant difference in performance. The average sum of squared torques is slightly above the prediction interval, but only by 1.16\% and including or not the value in the interval is approximately arbitrary. Consequently, the ablation of the pretraining phase does not have a noticeable effect on performance, and the pretraining phase can be removed from the pipeline to promote simplicity at 90\% certainty.

\section{Conclusion}
In this paper we proposed a competitive generalist locomotion policy for the ANYmal robot. Our approach achieves competitive results, on average about 90\% success rate at 95\% difficulty, on a set of experimental parkour trials of increasing complexity and difficulty.

To summarize, the mixture of experts approach currently leads in solving quadruped robots' agile locomotion challenges. However, history shows that inductive biases are suboptimal for long-term machine learning solutions.

Our ablation study highlights that every proprioception component is crucial for the end-result performance, while the history of proprioceptive states offers mixed results. Penalties on joint acceleration and stumbling improve behavior, but torque variation penalties are mostly redundant. Clipping standard deviation and batch sizes are vital for convergence, while pretraining appears inessential.

Future work includes refining design choices, such as training environments, standard deviation limits, and neural network architecture. Testing fully trained models and expanding the experiments for reproducibility are our next steps. Additionally, leveraging more computational power could improve competitiveness while transitioning to hardware implementation to validate real-world performance.

%
%
%

\newpage
\bibliographystyle{splncs04}
\bibliography{references}

\end{document}